\documentclass[10pt,twocolumn,letterpaper]{article}
\usepackage{cvpr} 
\usepackage{times}
\usepackage{epsfig}
\usepackage{graphicx}
\usepackage{amsmath}
\usepackage{amssymb}
\usepackage{booktabs}
\usepackage{xcolor}
\usepackage{multirow}
\usepackage{subcaption}
\usepackage{hyperref}

\makeatletter
\renewcommand\paragraph{\@startsection{paragraph}{4}{\z@}%
  {0.5em \@plus 0.2em \@minus 0.1em}%
  {-1em}%
  {\normalfont\normalsize\bfseries}}
\makeatother

\newcommand{\dbname}{DriveFace {}}
\newcommand{\dbnamepad}{DriveFace-PAD {}}

\begin{document}

\title{DriveFace: A Cross-Spectral Through-Glass Face Dataset for On-the-Move Vehicular Border Control}

\author{
Anjith George$^{1}$ \quad
Luis S. Luevano$^{1}$ \quad
Alain Komaty$^{1}$ \quad
Zeina Al Amine$^{1}$ \quad
Vidit Vidit$^{1}$ \quad
S\'ebastien Marcel$^{1,2}$\\
$^{1}$Idiap Research Institute, Rue Marconi 19, 1920 Martigny\\
$^{2}$University of Lausanne (UNIL), Lausanne, Switzerland\\
{\ttfamily\small
\begin{tabular}{@{}l@{}}
\{anjith.george, luis.luevano, alain.komaty,\\
zeina.alamine, vidit.vidit, sebastien.marcel\}@idiap.ch
\end{tabular}
}
}

\maketitle
\thispagestyle{empty}

\begin{abstract}
The continuous growth in cross-border mobility places increasing pressure on existing border control infrastructures, motivating \emph{on-the-move} biometric authentication, in which travellers are identified directly inside their vehicles at checkpoints. Face recognition is well-suited to this setting, as it can be acquired passively and at a distance. Its development, however, is hindered by the lack of representative datasets: existing benchmarks are collected in controlled environments and do not capture the challenges inherent to vehicular acquisition, including motion blur, variable illumination, occlusions, and cross-spectral enrollment. To address this gap, we introduce a dataset for on-the-move face recognition in border-control scenarios, comprising NIR vehicle-crossing videos paired with smartphone-based pre-enrollment data. Baseline evaluations with state-of-the-art models show clear performance limitations under these realistic conditions, highlighting the need for dedicated methods to advance the field.
\end{abstract}
\section{Introduction}
Face recognition \cite{kim202650} is a highly suitable biometric modality for automated border control (ABC) applications, owing to its non-contact acquisition capability and seamless integration with established workflows~\cite{labati2016biometric}. Nevertheless, achieving robust performance at land border crossings remains a significant challenge. Under operational conditions, subjects are frequently captured with non-frontal head pose, motion blur, variable ambient illumination, and partial facial occlusion. These difficulties are further exacerbated when image acquisition is performed through automotive windows, which introduces additional optical degradations that materially degrade recognition accuracy.

\begin{figure}[!b]
    \centering

    \begin{subfigure}[t]{0.31\linewidth}
        \centering
        \includegraphics[width=\linewidth]{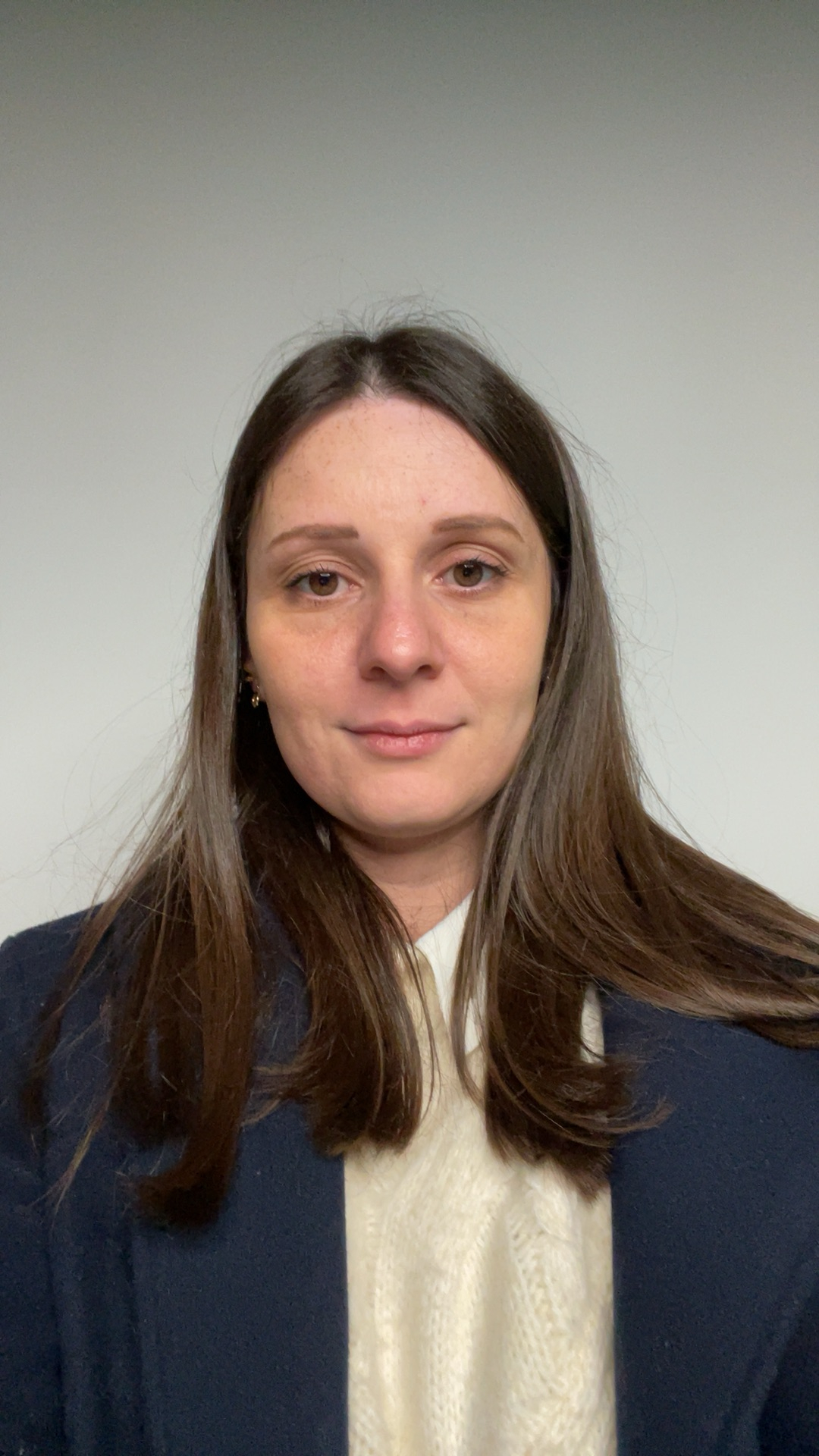}
        \caption{iPhone reference}
    \end{subfigure}\hfill
    \begin{subfigure}[t]{0.63\linewidth}
        \centering
        \includegraphics[width=\linewidth]{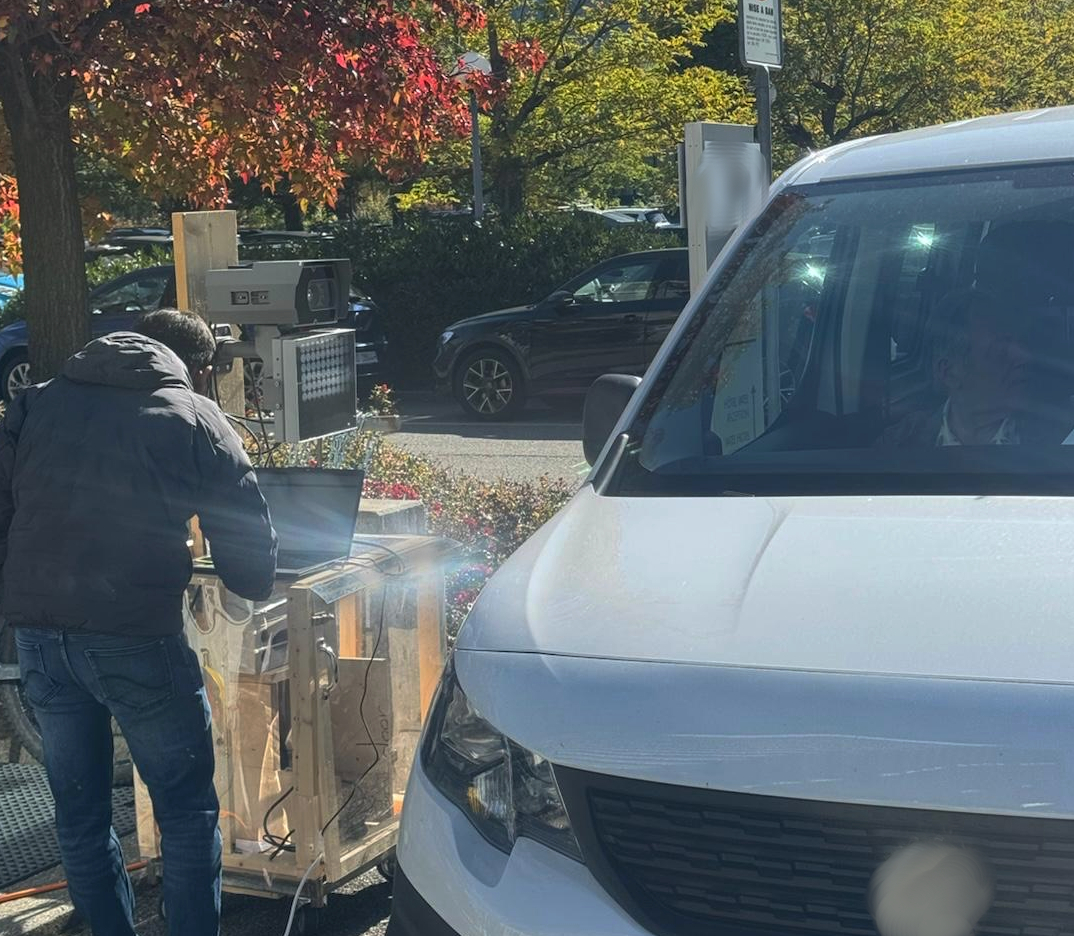}
        \caption{Sensor-vehicle outdoor setup (90°)}
    \end{subfigure}\hfill
    \vspace{0.25em}
    \begin{subfigure}[t]{0.31\linewidth}
        \centering
        \includegraphics[width=\linewidth]{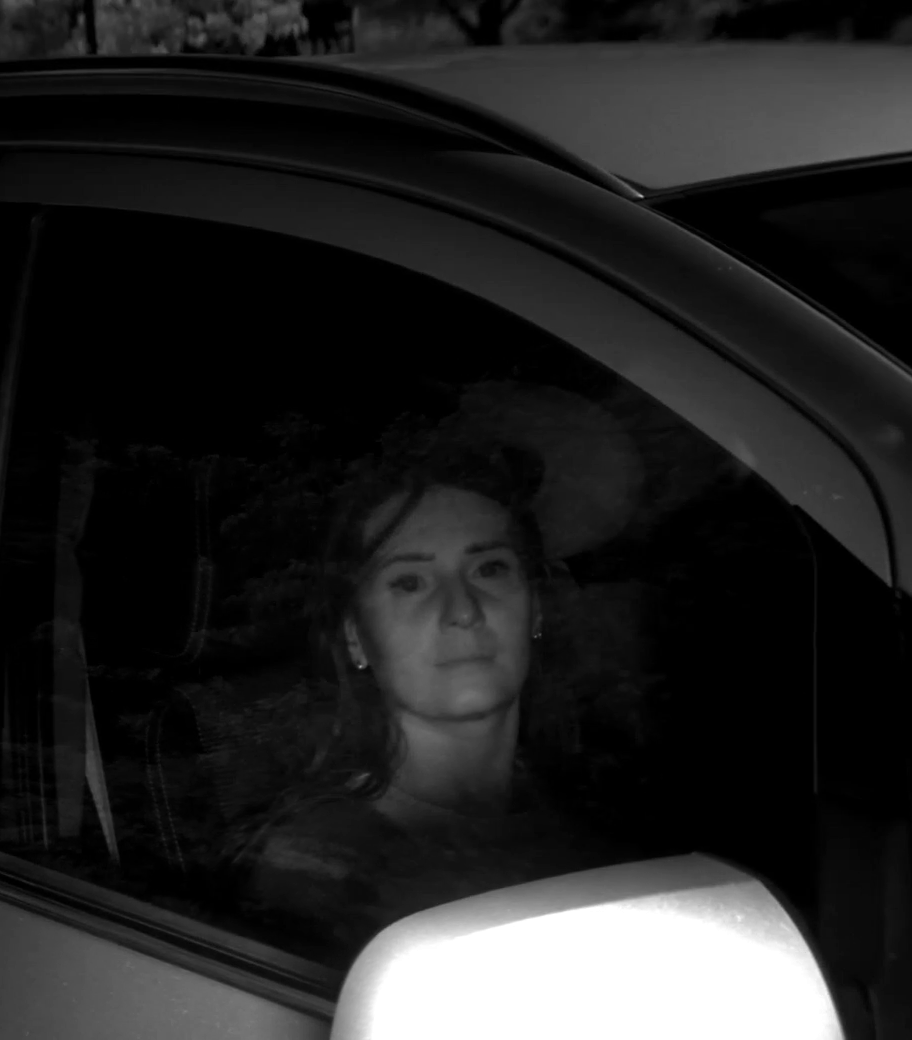}
        \caption{Outdoor (45°)}
    \end{subfigure}\hfill
    \begin{subfigure}[t]{0.31\linewidth}
        \centering
        \includegraphics[width=\linewidth]{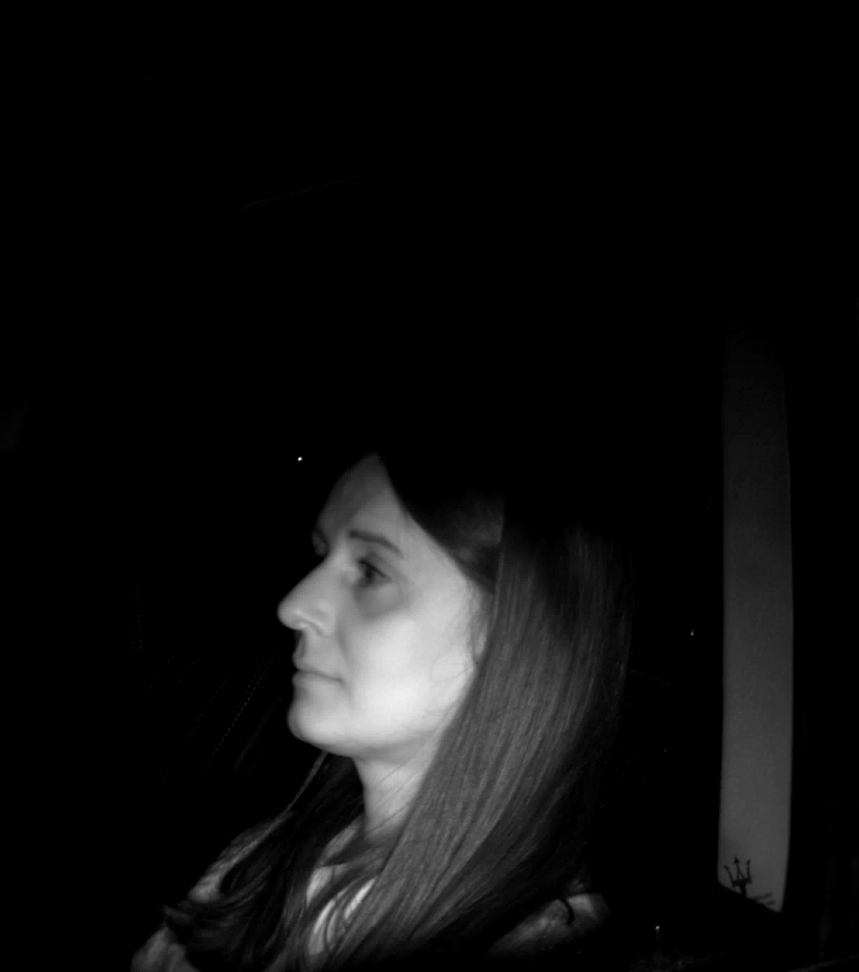}
        \caption{Indoor car}
    \end{subfigure}\hfill
    \begin{subfigure}[t]{0.31\linewidth}
        \centering
        \includegraphics[width=\linewidth]{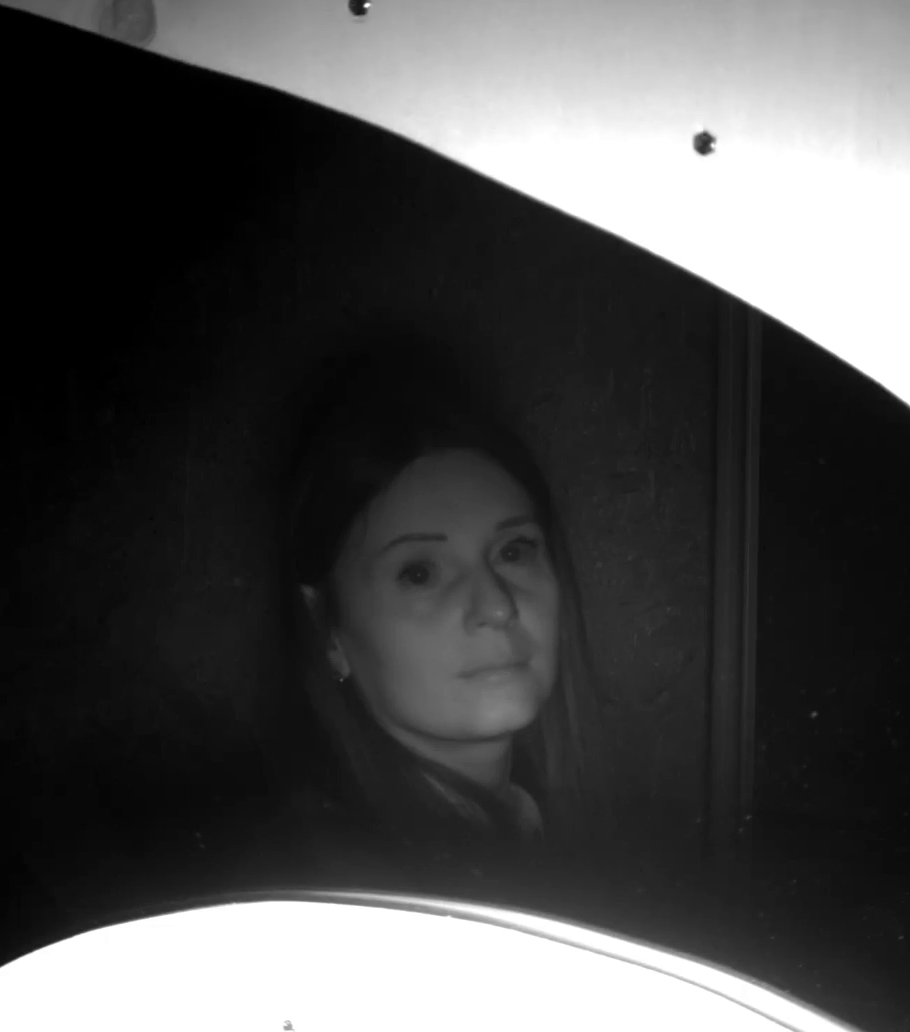}
        \caption{Simulated tinted window}
    \end{subfigure}

    \caption{Overview of the acquisition setup in \dbname{}, including a reference image, the probe VIDAR PAX \cite{adaptiverecognition_vidar_user_manual_2025} sensor-vehicle setup, and samples from the three main probe capture settings.}
    \label{fig:dataset_setup_overview}
\end{figure}

A particularly underexplored yet operationally significant scenario is \emph{in-vehicle border screening}, in which a traveler is observed while seated inside a vehicle through a windshield or side window. In addition to the typical challenges inherent to unconstrained face recognition, automotive windows introduce complex optical phenomena. Modern windshields frequently incorporate infrared-reflective coatings engineered to attenuate solar heat gain; such coatings can reflect in excess of 50\% of near-infrared (NIR) radiation, substantially reducing the signal available to NIR-based imaging systems. Variability in window tint across vehicle makes and models further increases variability ~\cite{kuchar2023effect}, while specular reflections arising from ambient lighting introduce additional noise and artifacts.

VIS-NIR face recognition further suffers from the cross-spectral domain gap~\cite{anghelone2025beyond}: enrollment references are typically acquired in the visible (VIS) spectrum (\eg, passport or mobile photographs), whereas checkpoint probes are captured in NIR, yielding substantial differences in facial texture and reflectance that challenge models trained on VIS data alone. Although benchmarks such as CASIA NIR-VIS 2.0~\cite{li2013casia} have advanced heterogeneous face recognition, they are collected in controlled settings and do not reproduce the combined effects of through-glass imaging, subject motion, and unconstrained outdoor environments. Datasets acquired in operational border settings remain largely proprietary~\cite{cornett2018through}.

To address these limitations, we introduce \dbname{}, a NIR dataset that captures the end-to-end biometric workflow at vehicular border crossings and enables systematic study of cross-spectral matching, through-glass degradation, and adverse illumination within a single operationally representative benchmark collected over two months. A dedicated presentation attack subset further supports anti-spoofing research under realistic deployment conditions. Figure~\ref{fig:dataset_setup_overview} illustrates the capture settings.

The main contributions of this work are listed below:
\begin{itemize}
    \item \dbname{}, a publicly available dataset for face recognition at vehicular border crossings, pairing VIS pre-enrollment images with in-vehicle NIR probes acquired through automotive windows, and a presentation attack subset (\dbnamepad{}).
    \item Rich per-acquisition metadata: tint level, illumination, head pose, and vehicle speed, enabling fine-grained analysis of individual degradation sources.
    \item Standardized face recognition and PAD protocols with baseline results establishing reference performance on \dbname{}. The dataset is available at: \footnote{\url{https://www.idiap.ch/paper/driveface}
    }
    \footnote{\url{https://www.idiap.ch/en/scientific-research/data/driveface}}.
\end{itemize}

\section{Related Work}
While there are a large number of face recognition datasets publicly available~\cite{pasc,msceleb1m,arcface-ms1mv2,lfrc-iccvw19-ms1mv3,zhu2021webface260m}, most are collected in controlled environments and do not reflect the challenges of vehicular border control. Below, we review datasets relevant to this setting.
\paragraph{In-Vehicle Face Datasets}
In-vehicle face datasets are typically introduced for driver monitoring rather than identity recognition. Representative examples include Drive\&Act~\cite{drive_and_act_2019_iccv}, DMD~\cite{dmd}, and DriverMVT~\cite{drivermvt}, which capture in-cabin subjects under varying pose, occlusion, illumination, and seating position for tasks such as distraction, drowsiness, gaze, and head-pose estimation. Datasets more oriented toward in-vehicle authentication include AVICAR~\cite{avicar}, VFPAD~\cite{vfpad}, and iCarB~\cite{icarb}. While valuable for automotive facial analysis, none target heterogeneous cross-session identity matching from \emph{outside} the vehicle through tinted or clear windows with profile-pose variation, leaving a clear gap for external-view, cross-modal recognition.
\paragraph{NIR Face Datasets}
Near-infrared heterogeneous datasets support face recognition beyond the visible spectrum, particularly under low light and for RGB-to-NIR matching. Representative datasets include CASIA NIR-VIS 2.0~\cite{li2013casia}, PolyU-NIRFD~\cite{polyu-nirfd}, Oulu-CASIA NIR\&VIS~\cite{oulu-casia-nir-vis}, and BUAA-VisNir~\cite{buaa-visnir}, alongside broader multimodal datasets such as the Tufts Face Database~\cite{tufts} and LAMP-HQ~\cite{lamp-hq}. Although these have advanced heterogeneous face recognition, they are collected indoors under controlled conditions and do not capture through-glass acquisition, in-vehicle deployment, motion, or outdoor settings.
\paragraph{Face Acquisition Through Glass}
Through-glass face acquisition remains relatively underexplored. The most related works are the through-windshield driver recognition study of Cornett~\etal~\cite{cornett2018through}, the Face Image Reflection Removal (FIRR) dataset~\cite{face-image-reflection-removal}, and PP4AV~\cite{pp4av}, which cover subjects viewed through windshields, real outdoor driving, and reflection-degraded imagery, respectively. However, \cite{face-image-reflection-removal} and \cite{pp4av} target different tasks and do not model heterogeneous mobile-RGB to external-NIR matching, while \cite{cornett2018through} captures multiple occupants through tinted windows. Public benchmarks for through-glass face recognition therefore remain limited.
\paragraph{Presentation Attack Detection Datasets}
PAD datasets assess robustness to print, replay, and mask attacks. Established benchmarks such as Replay-Attack~\cite{replay-attack-idiap}, CASIA-FASD~\cite{casia-fasd}, MSU-MFSD~\cite{msu-mfsd}, OULU-NPU~\cite{oulu-npu}, and SiW~\cite{siw-2018} rely on RGB capture in mobile or desktop settings, while more recent datasets such as CASIA-SURF~\cite{casia-surf}, CASIA-SURF CeFA~\cite{casia-surf-cefa}, CelebA-Spoof~\cite{celeba-spoof}, WMCA~\cite{wmca}, and HiFiMask~\cite{hifimask} extend toward multi-modal sensing, cross-ethnicity evaluation, and 3D masks. However, none of them jointly address PAD in NIR, through-glass acquisition, and external-view capture under tint, reflections, and motion. The in-vehicle datasets VFPAD~\cite{vfpad} and iCarB~\cite{icarb} are closer: VFPAD targets in-vehicle NIR PAD and iCarB supports in-car PAD-oriented evaluation, but both focus on in-cabin capture rather than external probes acquired through automotive glass.

\section{The DriveFace Dataset}
\begin{figure}[t]
    \centering

    \begin{minipage}[c]{0.07\linewidth}
        \centering
        \rotatebox{90}{\small \textbf{Reference}}
    \end{minipage}%
    \begin{minipage}[c]{0.91\linewidth}
        \centering
        \begin{subfigure}[t]{0.455\linewidth}
            \centering
            \includegraphics[width=\linewidth,height=3cm,keepaspectratio]{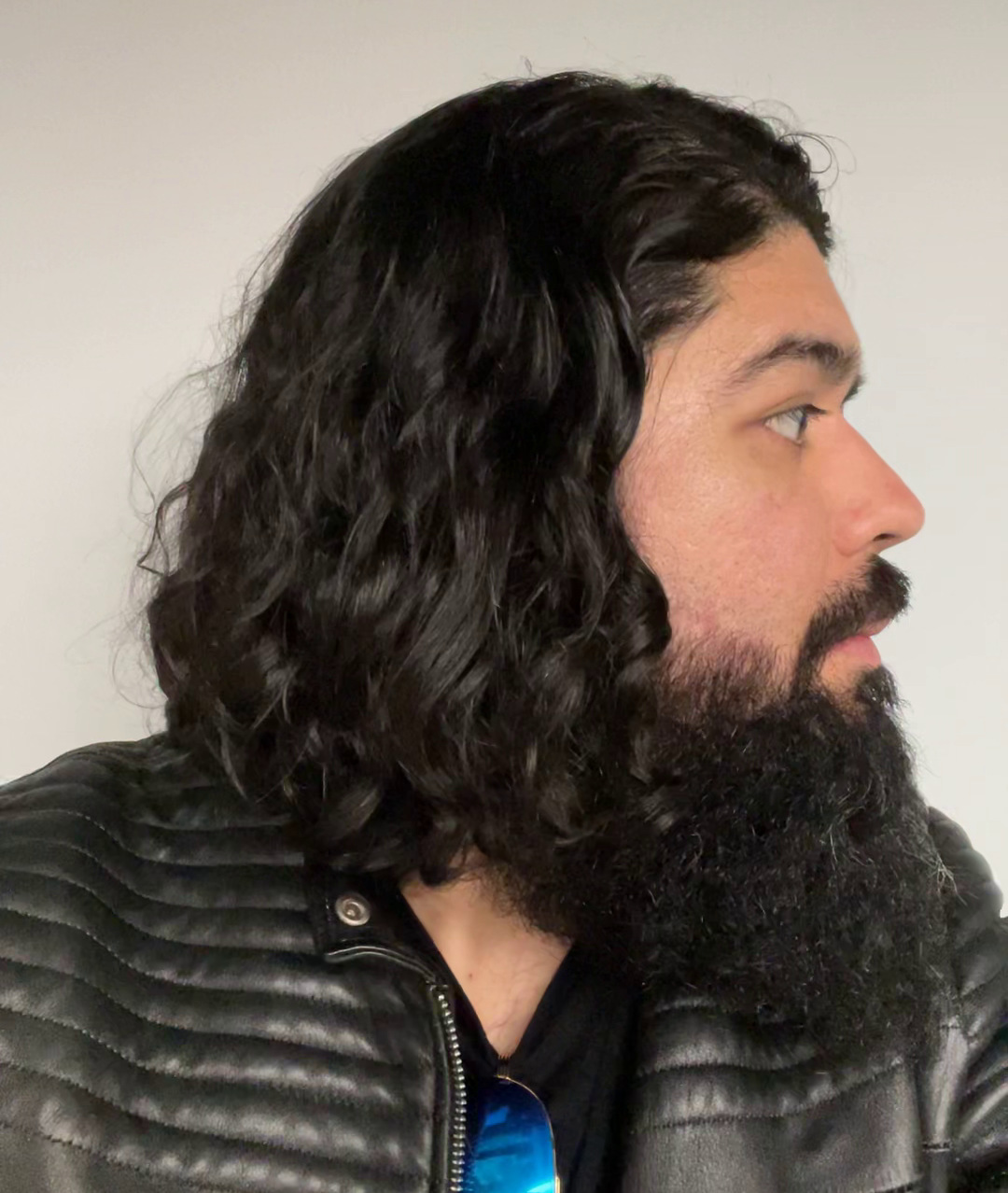}
            \caption{Session 1, hard profile}
        \end{subfigure}\hspace{0.012\linewidth}%
        \begin{subfigure}[t]{0.455\linewidth}
            \centering
            \includegraphics[width=\linewidth,height=3cm,keepaspectratio]{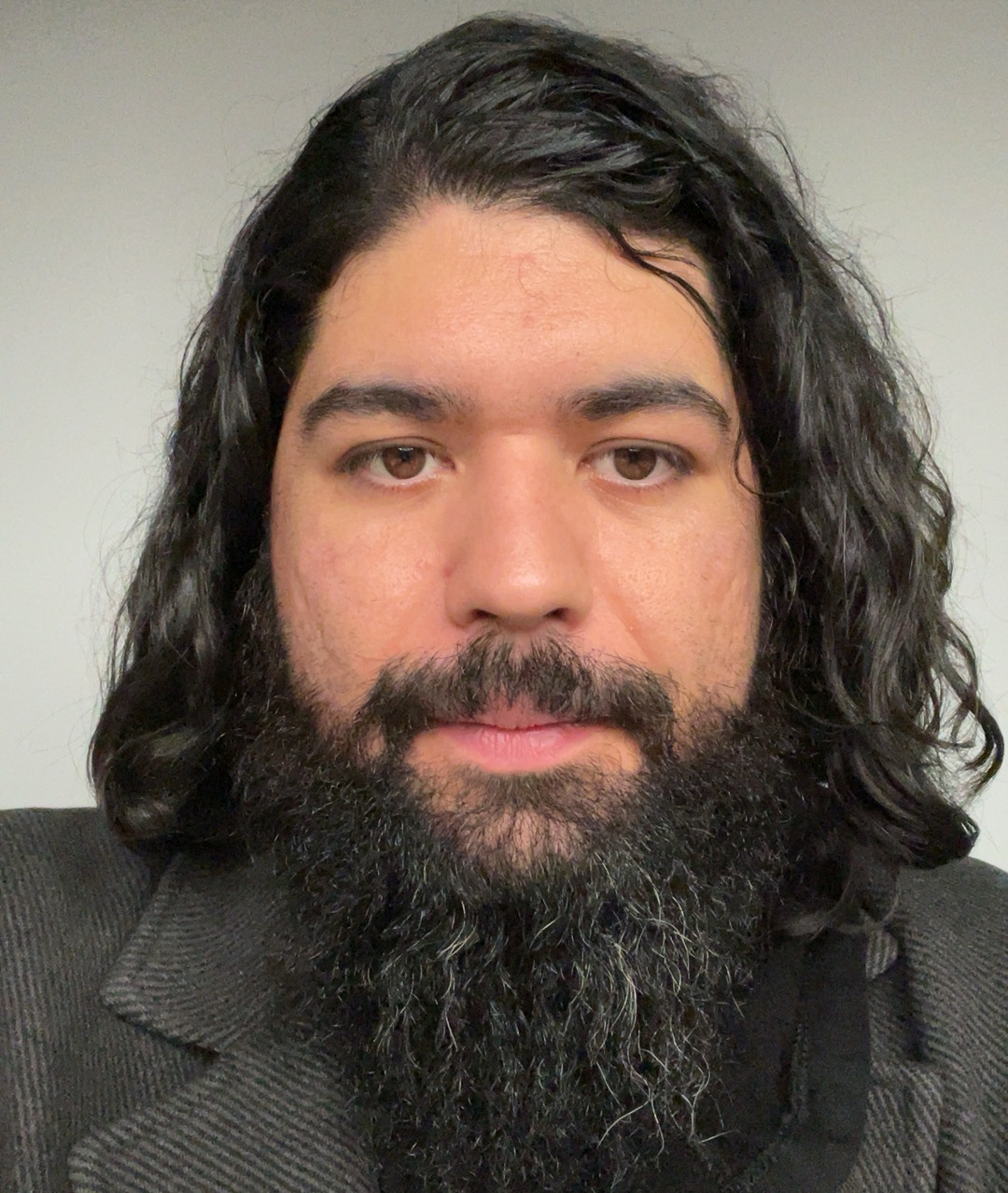}
            \caption{Session 2, frontal}
        \end{subfigure}
    \end{minipage}
    \vspace{0.2em}
    \begin{minipage}[c]{0.07\linewidth}
        \centering
        \rotatebox{90}{\small \textbf{Outdoor}}
    \end{minipage}%
    \begin{minipage}[c]{0.91\linewidth}
        \centering
        \begin{subfigure}[t]{0.455\linewidth}
            \centering
            \includegraphics[width=\linewidth,height=3cm,keepaspectratio]{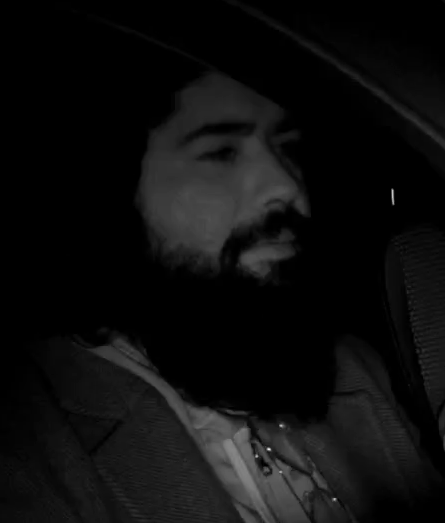}
            \caption{Clear window, $45^\circ$, moving}
        \end{subfigure}\hspace{0.012\linewidth}%
        \begin{subfigure}[t]{0.455\linewidth}
            \centering
            \includegraphics[width=\linewidth,height=3cm,keepaspectratio]{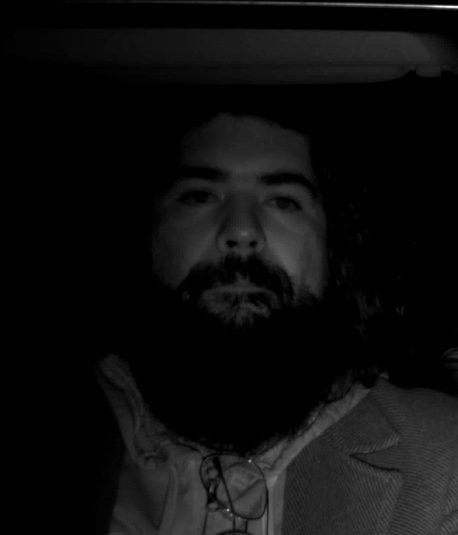}
            \caption{Clear window, $0^\circ$, stationary}
        \end{subfigure}
    \end{minipage}
    \vspace{0.2em}
    \begin{minipage}[c]{0.07\linewidth}
        \centering
        \rotatebox{90}{\small \textbf{Sim. tint indoor}}
    \end{minipage}%
    \begin{minipage}[c]{0.91\linewidth}
        \centering
        \begin{subfigure}[t]{0.303\linewidth}
            \centering
            \includegraphics[width=\linewidth,height=3cm,keepaspectratio]{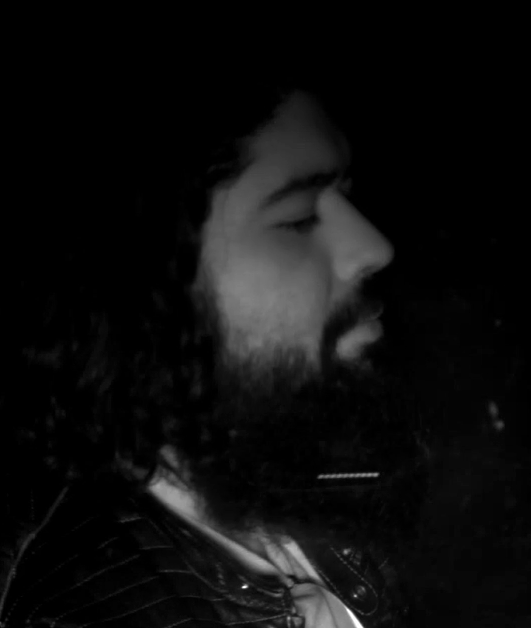}
            \caption{Clear window}
        \end{subfigure}\hspace{0.008\linewidth}%
        \begin{subfigure}[t]{0.303\linewidth}
            \centering
            \includegraphics[width=\linewidth,height=3cm,keepaspectratio]{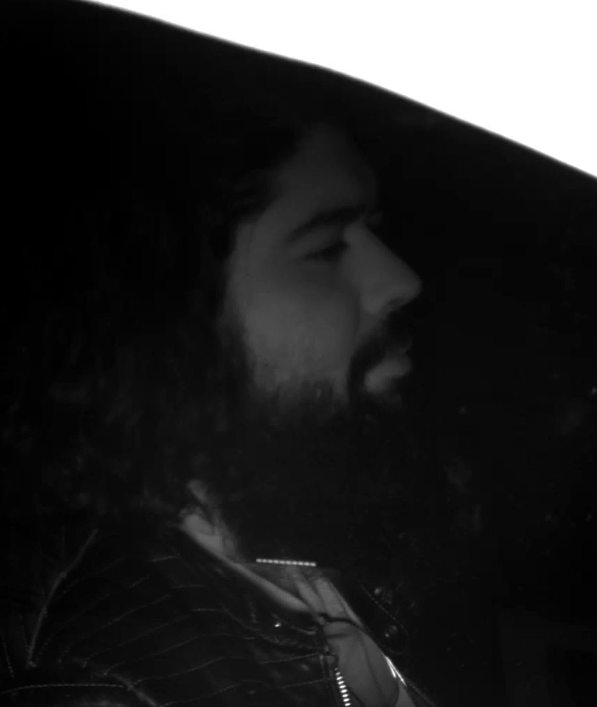}
            \caption{Med. tint (T35)}
        \end{subfigure}\hspace{0.008\linewidth}%
        \begin{subfigure}[t]{0.303\linewidth}
            \centering
            \includegraphics[width=\linewidth,height=3cm,keepaspectratio]{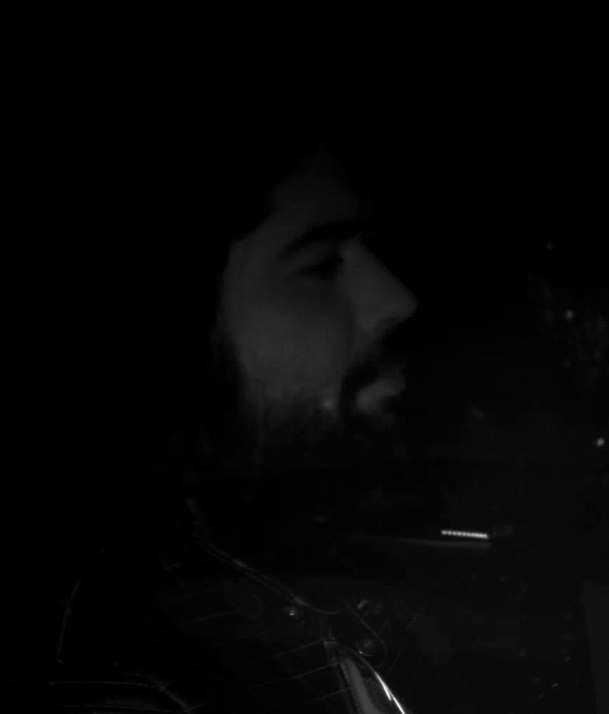}
            \caption{Dark tint (T05)}
        \end{subfigure}
    \end{minipage}

    \caption{Samples from a single subject in \dbname{}. The figure shows visible-spectrum reference captures from two sessions, infrared probe samples acquired through simulated indoor window conditions with different tint levels and real-world outdoor infrared acquired through clear vehicle windows at different viewing angles and moving conditions.}
    \label{fig:new_dataset_samples}
\end{figure}

The proposed dataset was designed to support biometric research in operational border-control settings. It consists of three components: \emph{(i) pre-enrollment face captures}, \emph{(ii) in-vehicle face captures}, and \emph{(iii) presentation attacks}. Together, these components reflect a realistic workflow in which a traveler first provides reference biometric data prior to arrival and is subsequently observed under challenging checkpoint conditions. The dataset was collected from \textbf{70 consenting subjects} over \textbf{two recording sessions}, with corresponding reference samples and probes. Figure \ref{fig:new_dataset_samples} shows examples of the pre-enrollment face captures for reference and probe images. For most participants, the two sessions were collected approximately two months apart in order to capture realistic short-term variation in appearance, including changes in facial hair, hairstyle, grooming, makeup, and other mild changes in facial appearance.

\begin{table}
\centering
\caption{Summary of the main components of \dbname{}.}
\label{tab:dataset_summary}
\small
\setlength{\tabcolsep}{4pt}
\begin{tabular}{lccc}
\toprule
\textbf{Subset} & \textbf{Mod.} & \textbf{\# Var.} & \textbf{Face pose} \\
\midrule
Reference        & RGB  & 4  & frontal to profile  \\
Outdoor probes   & NIR  & 24  & frontal, 3/4, profile \\
Sim. tint indoor & NIR  & 36 & frontal, profile \\
Indoor car       & NIR   & 6 & frontal, profile \\
\bottomrule
\end{tabular}
\end{table}

\begin{figure}[tb]
    \centering
    \begin{minipage}[c]{0.07\linewidth}
        \centering
        \rotatebox{90}{\small \textbf{Indoor}}
    \end{minipage}\hfill
    \begin{minipage}[c]{0.91\linewidth}
        \centering
        \begin{subfigure}[t]{0.31\linewidth}
            \centering
            \includegraphics[width=\linewidth]{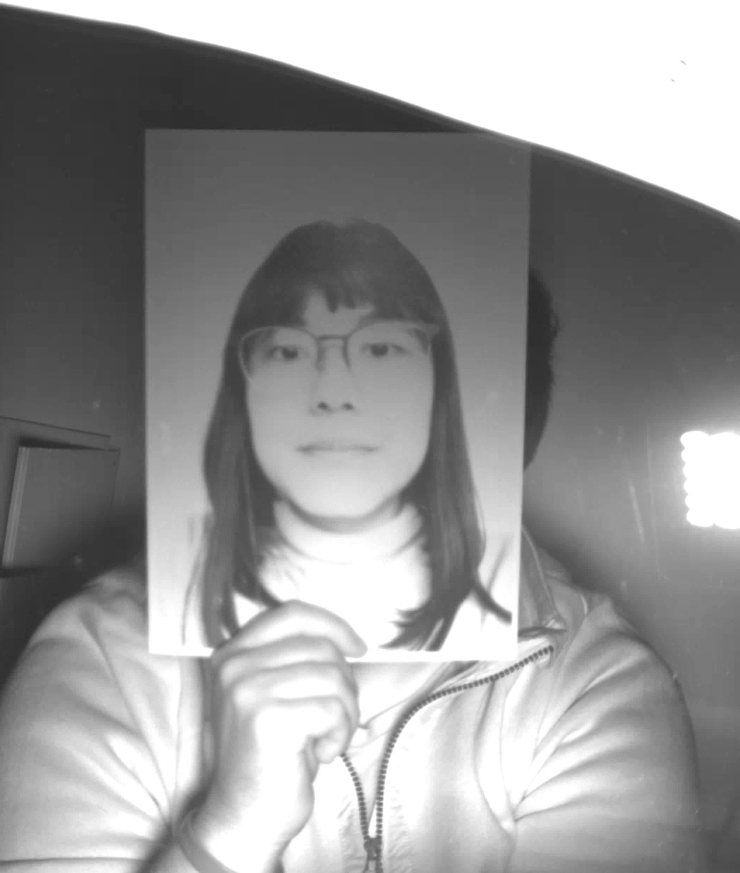}
            \caption{Laser-Matte print, medium tint (T20)}
        \end{subfigure}\hfill
        \begin{subfigure}[t]{0.31\linewidth}
            \centering
            \includegraphics[width=\linewidth]{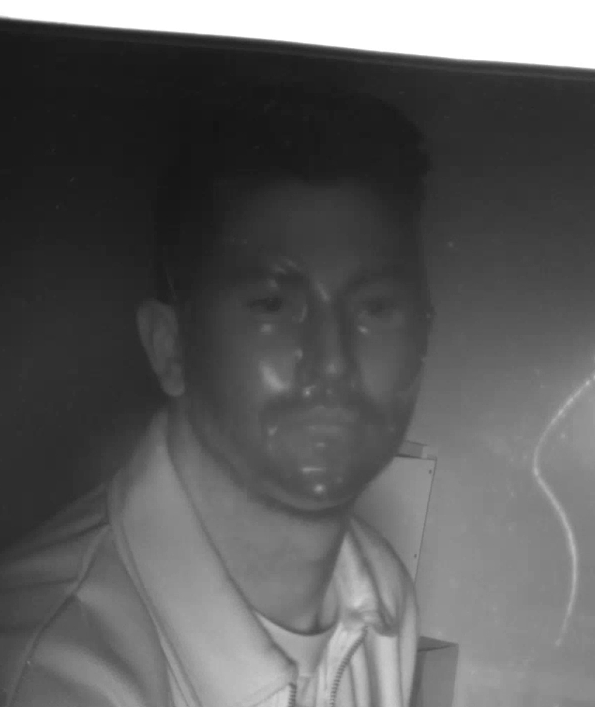}
            \caption{Transparent mask, dark tint (T05)}
        \end{subfigure}\hfill
        \begin{subfigure}[t]{0.31\linewidth}
            \centering
            \includegraphics[width=\linewidth]{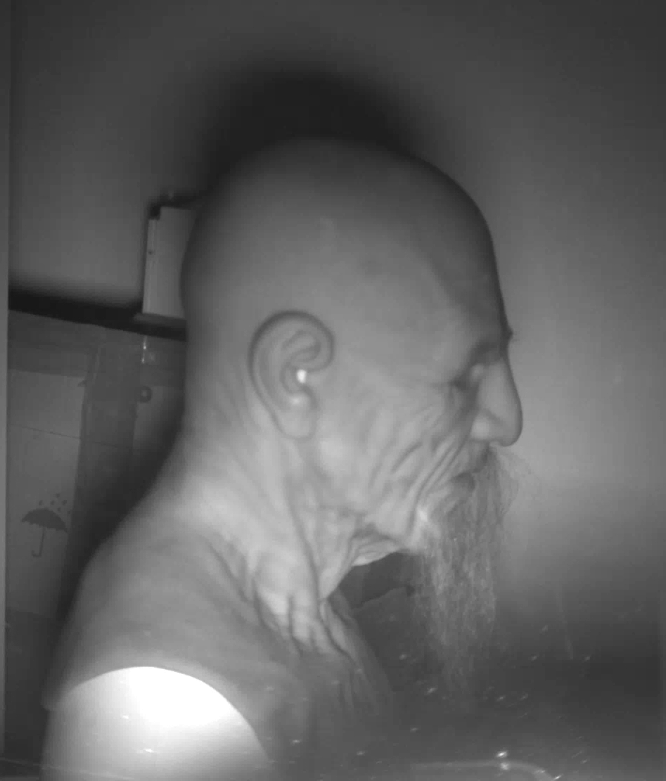}
            \caption{Silicone-mannequin, medium tint (T20)}
        \end{subfigure}
    \end{minipage}
    \vspace{0.4em}
    \begin{minipage}[c]{0.07\linewidth}
        \centering
        \rotatebox{90}{\small \textbf{Outdoor}}
    \end{minipage}\hfill
    \begin{minipage}[c]{0.91\linewidth}
        \centering
        \begin{subfigure}[t]{0.31\linewidth}
            \centering
            \includegraphics[width=\linewidth]{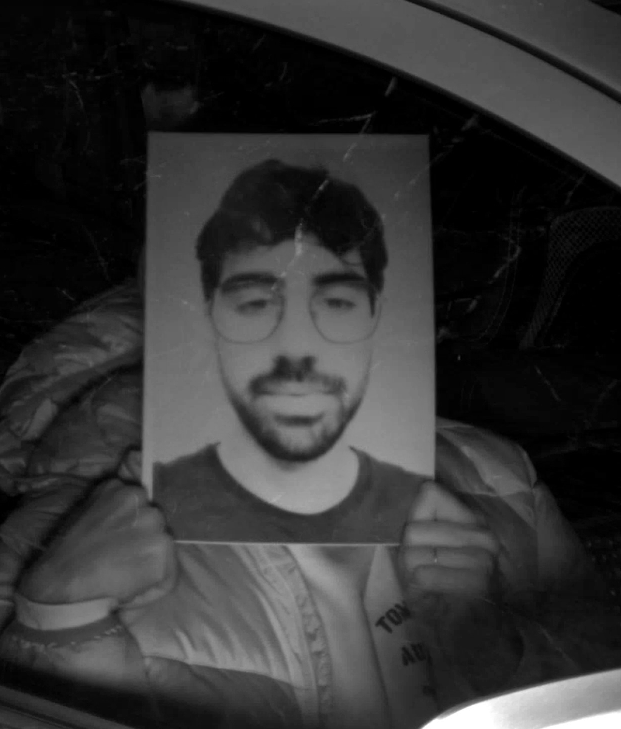}
            \caption{Laser-Glossy, frontal}
        \end{subfigure}\hfill
        \begin{subfigure}[t]{0.31\linewidth}
            \centering
            \includegraphics[width=\linewidth]{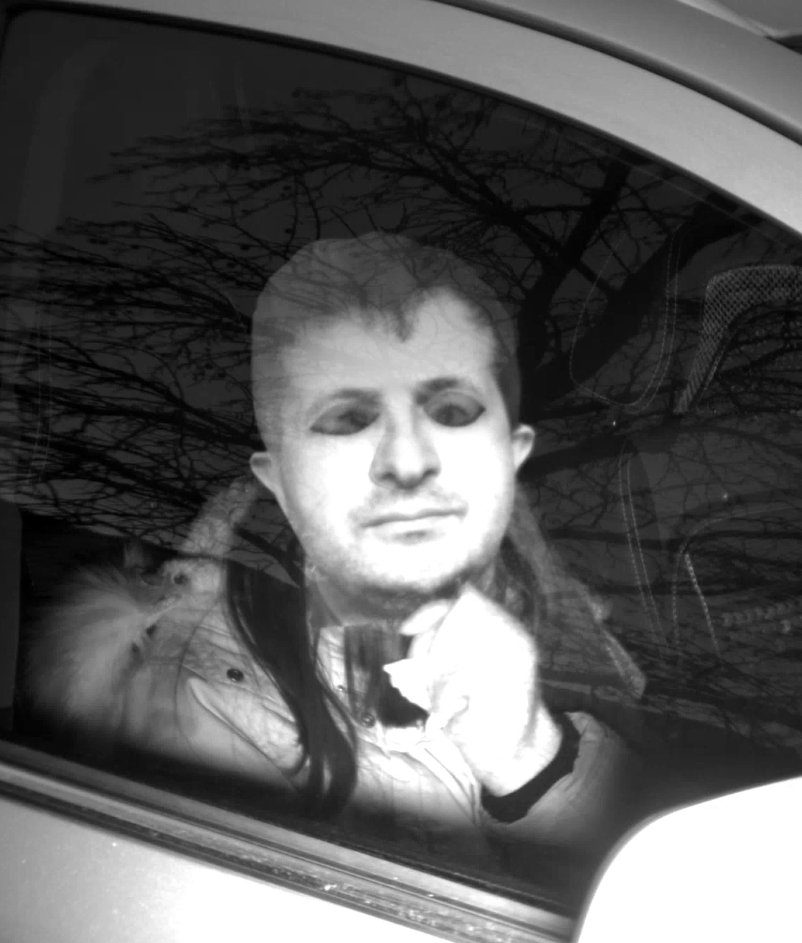}
            \caption{Paper mask, frontal}
        \end{subfigure}\hfill
        \begin{subfigure}[t]{0.31\linewidth}
            \centering
            \includegraphics[width=\linewidth]{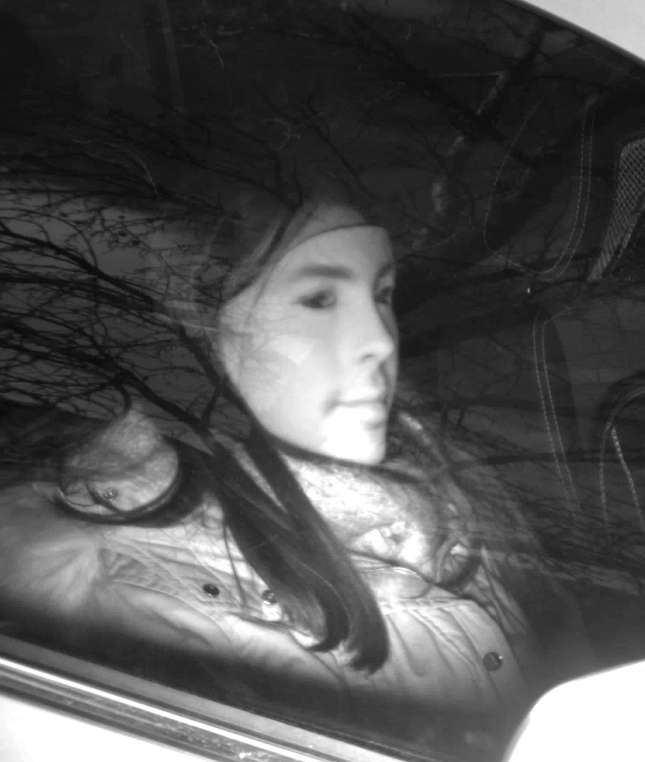}
            \caption{Resin mask, profile}
        \end{subfigure}
    \end{minipage}

    \caption{Example presentation attack samples from \dbnamepad{}. The figure shows variation in attack types under indoor and outdoor conditions, with examples covering different presentation poses and levels of glass tint.}
    \label{fig:pad_samples}
\end{figure}

\begin{table}[t]
    \centering
    \caption{Summary of the presentation attack subset in \dbnamepad{}.}
    \label{tab:pad_dataset_summary}
    \resizebox{\linewidth}{!}{%
        \begin{tabular}{l c p{4.6cm}}
            \toprule
            \textbf{Attack} &
            \textbf{Identity} &
            \textbf{Instruments / materials} \\
            \midrule
            Print &
            Bona fide IDs &
            \begin{tabular}[t]{@{}l@{}}
                Printer: Inkjet / Laser \\
                Matte / Normal / Glossy
            \end{tabular} \\
            Replay &
            Bona fide IDs &
            Laptop display \\
            Mask &
            External IDs &
            \begin{tabular}[t]{@{}l@{}}
                Paper, resin, silicone, \\
                plastic, plastic + makeup, \\
                thin colored plastic, rubber, \\
                silicone, mannequin
            \end{tabular} \\
            \bottomrule
        \end{tabular}%
    }
\end{table}

\subsection{Demographics}
\begin{figure}
    \centering
    \includegraphics[width=1\linewidth]{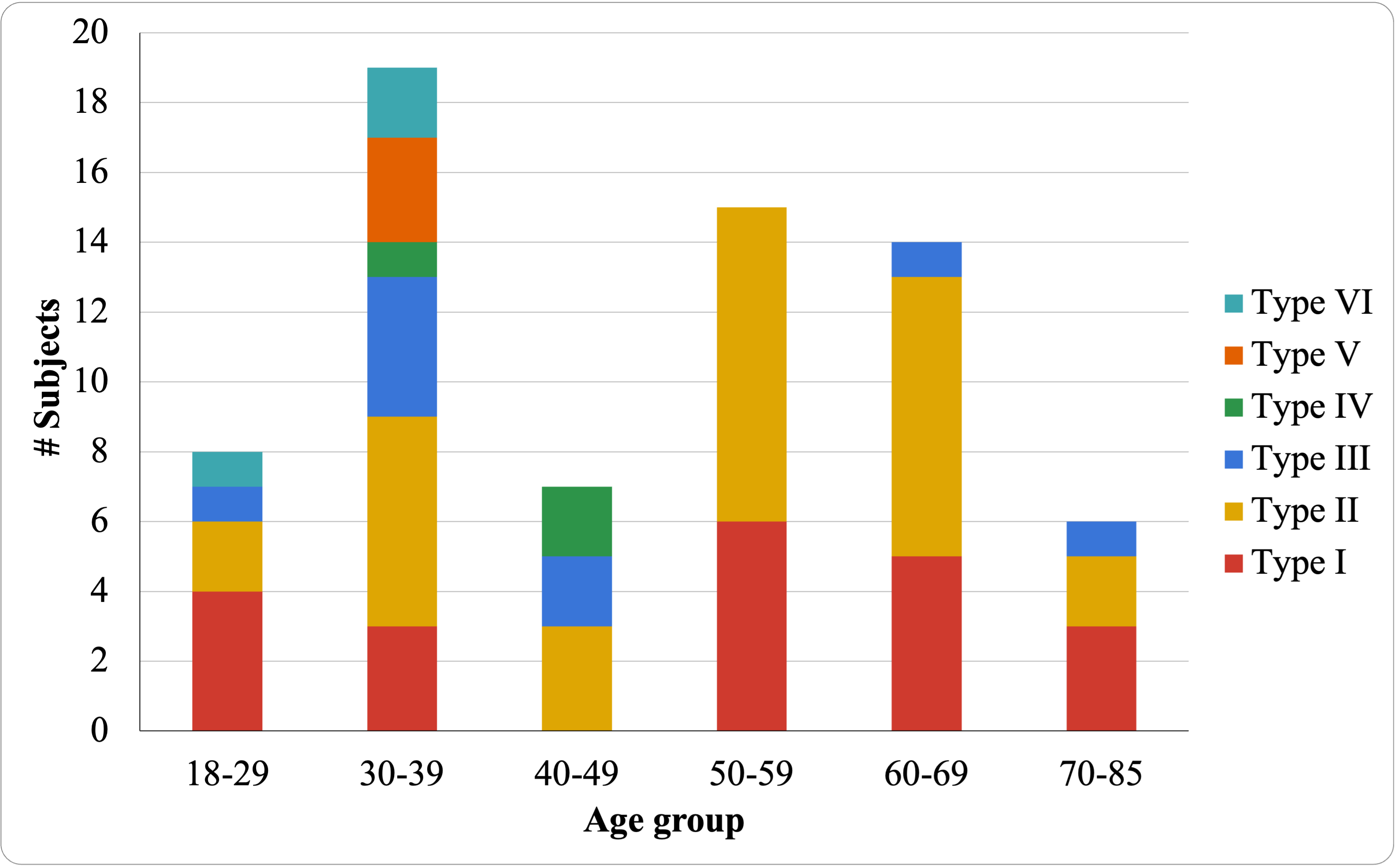}
    \caption{Age group distribution labeled by Fitzpatrick skin color tones.}
    \label{fig:age_skin}
\end{figure}

The dataset was collected from \emph{70 consenting volunteers}. The same participant pool was used across the data collection campaign, which allows direct association between pre-enrollment reference samples and in-vehicle operational captures. This design supports both verification and identification experiments in which clean enrollment data are matched against challenging probes. This dataset effort contains a demographically diverse group of participants. In terms of gender, the dataset is relatively balanced, with 45.7\% female and 54.3\% male subjects. The age distribution spans a broad range from 18 to 85 years old. The dataset also includes variation in apparent skin tone, which we summarize into three broad groups: light, medium, and dark, containing 51, 11, and 6 subjects, respectively, based on the available annotations. Figure \ref{fig:age_skin} illustrates the age and skin-tone distributions of the subjects in the dataset.

\subsection{Pre-enrollment Face Captures}
The pre-enrollment subset was designed to simulate a trusted-traveler or pre-registration procedure performed before arrival at a checkpoint. For each subject and each session, close-range facial videos were acquired using the front-facing cameras of two consumer smartphones: an \emph{iPhone 12} and a \emph{Samsung Galaxy S9}. This subset provides high-quality reference data intended to serve as identity ground truth for subsequent comparisons with more challenging operational captures. During acquisition, subjects were instructed to perform controlled head movements in order to introduce natural pose variation rather than a single static frontal view. This subset is therefore suitable as an enrollment/reference set for verification and identification experiments under cross-device and cross-condition settings.
\subsection{In-vehicle Infrared Face Captures}
The in-vehicle subset targets face acquisition and recognition for subjects seated inside vehicles and observed through automotive glass. This scenario represents a particularly challenging condition for border biometrics, due to the combined effects of reflections, window tint, low transmission, viewpoint changes, motion blur, and reduced illumination. The in-vehicle data were acquired primarily in the \emph{near-infrared (NIR)} spectrum under two complementary protocols using a VIDAR PAX infrared sensor \cite{adaptiverecognition_vidar_user_manual_2025} with an external illuminator.\\
\textbf{Real-vehicle outdoor captures:} In the first protocol, subjects were recorded inside a real vehicle from an acquisition distance of approximately \emph{1--2\,m} from the glass. Recordings were captured from multiple viewpoints, including approximately \emph{$0^\circ$}, \emph{$45^\circ$}, and \emph{$90^\circ$}, and under both \emph{stationary} and \emph{moving-vehicle} conditions. This protocol was designed to reproduce realistic checkpoint approach conditions, including motion-related degradation and perspective distortion.\\
\textbf{Controlled indoor simulated-car captures:} In the second protocol, a controlled indoor setup was used to isolate the effects of glass transmission and scene illumination. Recordings were acquired under \emph{five window tint levels} corresponding to \emph{5, 15, 20, 30, and 45 Visible Light Transmission (VLT) levels}, and under \emph{three illumination levels} of \emph{5\%}, \emph{20\%}, and \emph{30\%}. Instead of using real cars, we mounted automotive glass panels with different VLT levels on wooden frames simulating these variations. This protocol provides a structured benchmark for studying the limits of face recognition systems under through-glass conditions.\\
\textbf{Real-vehicle indoor captures:} For the third protocol, a smaller set of real-vehicle indoor captures was also collected to study face acquisition in a more controlled environment while preserving the in-car setting. In this protocol, subjects were recorded inside a real vehicle at approximately \emph{$90^\circ$} relative to the vehicle direction through the \emph{clear window}. Two facial pose conditions were considered, namely \emph{CAM} and \emph{FWD}, and recordings were acquired at \emph{5\%}, \emph{20\%}, and \emph{30\%} illuminator power. This part of the dataset contains \emph{418 NIR video clips}.\\
Overall, the in-vehicle subset provides a challenging and structured benchmark for evaluating robustness to through-glass degradation, viewpoint variation, and adverse lighting in border-control scenarios.
\subsection{Presentation Attacks}
The dataset also includes a presentation attack subset designed to evaluate the robustness of face anti-spoofing systems in the proposed vehicular border-control setting. Attacks are presented in outdoor and indoor conditions from inside the vehicle and captured from the outside using the infrared sensor positioned at approximately \emph{$45^\circ$} relative to the vehicle direction.

The PAD subset includes three attack categories: \emph{print}, \emph{replay}, and \emph{mask}. Print and replay attacks are generated from subjects for whom bona fide samples are available in other parts of the dataset, while mask attacks are collected from external subjects. All attack types are recorded under the same factors: \emph{four infrared illumination levels} (\emph{5\%, 10\%, 20\%, 30\%}) and \emph{two attack poses}, namely \emph{CAM} and \emph{FWD}. The subset also includes variation within each attack category. For \emph{print attacks}, the dataset contains different combinations of \emph{printer type} and \emph{paper material}, including \emph{inkjet} and \emph{laser} printers, and \emph{matte}, \emph{normal}, and \emph{glossy} paper. For replay attacks, it contains two \emph{laptop display} types. For \emph{mask attacks}, the dataset includes \emph{paper}, \emph{resin}, \emph{silicone}, \emph{transparent plastic}, \emph{transparent plastic with makeup}, \emph{thin colored plastic}, \emph{rubber}, \emph{silicone over mannequin}, and \emph{mannequin}. Figure~\ref{fig:pad_samples} shows example attacks, while Table~\ref{tab:pad_dataset_summary} summarizes the main factors covered by the subset.

\subsection{Annotations and Metadata}
Automated detection and tracking were used to isolate consenting participants and remove bystanders, followed by a final \emph{human-in-the-loop} check to ensure that no non-consenting identities remained visible. The dataset provides annotations and metadata at the \emph{subject}, \emph{session}, and \emph{video} levels, with most acquisition settings encoded in the filename. \emph{Identity labels} include the subject ID and, for in-vehicle captures, the identities and seat positions of all occupants. \emph{Session labels} are defined for both reference and probe data: reference filenames encode subject ID, date, and session number, while the parent folder indicates the mobile device; in the probe subset, session 1 denotes outdoor captures and session 2 indoor captures. Additional \emph{subject-level} metadata include age, gender, and skin tone. \emph{Viewpoint labels} specify camera direction relative to the vehicle (\emph{$0^\circ$}, \emph{$45^\circ$}, \emph{$90^\circ$}) and face pose (\emph{CAM}, \emph{FWD}). \emph{Tint and illumination metadata} include illuminator setting, scenario, speed, weather, and window tint level. For the PAD subset, \emph{attack labels} also encode the attack category, instrument, and material.

\section{Experiments }

This section describes the benchmarking performed on the \dbname dataset, including both FR and PAD experiments.

\subsection{Face Recognition Experiments}
\paragraph{Protocols}
In the operational setting, RGB videos are used for enrollment. These enrollment videos capture subjects performing controlled head movements (left–right and up–down). In the probe scenario, subjects are recorded using an NIR camera while seated inside vehicles under different operational conditions. For enrollment, a frontal face image is selected for each subject using a pose estimator. For the probe set, we use face crops produced by a face detector and manually annotated with identity labels corresponding to the operational scenarios, selecting up to 10 samples per subject from each video.

For training and evaluation, the dataset is partitioned into disjoint train and test sets based on identity. Specifically, 60\% of identities are used for training and the remaining 40\% for testing. Each subject has two capture sessions, where samples from different sessions are used for enrollment and probe sets to ensure cross-session evaluation.

The dataset includes three operational conditions: (1) outdoor in-vehicle scenarios, (2) indoor in-vehicle scenarios, and (3) indoor simulation scenarios with varying automotive glass tint levels. These correspond to the protocols \textit{outdoor}, \textit{indoor\_car}, and \textit{simulation}. Additional metadata, such as vehicle speed, tint level, weather, and illumination conditions are provided to support more detailed analysis. The dataset statistics are shown in Table \ref{tab:fr_db_stats}.

\begin{table}[t]
\centering
\caption{Face recognition protocol statistics for the \dbname\ dataset.}
\label{tab:fr_db_stats}
\resizebox{0.49\textwidth}{!}{%
\begin{tabular}{lccc ccc}
\toprule
& \multicolumn{3}{c}{\textbf{Train}} & \multicolumn{3}{c}{\textbf{Test}} \\
\cmidrule(lr){2-4} \cmidrule(lr){5-7}
\textbf{Protocol} & \textbf{IDs} & \textbf{Videos} & \textbf{Frames} & \textbf{IDs} & \textbf{Videos} & \textbf{Frames} \\
\midrule
outdoor     & 41 & 1178 & 11699 & 28 & 909 & 9057 \\
indoor\_car & 42 & 245  & 2050  & 28 & 166 & 1410 \\
simulation  & 42 & 1450 & 14100 & 28 & 928 & 8980 \\
\bottomrule
\end{tabular}
}
\end{table}

\paragraph{Metrics}
To evaluate face recognition performance, we report several standard metrics commonly used in the literature, including Area Under the Curve (AUC), Equal Error Rate (EER), Rank-1 identification rate, and Verification Rates (VR a.k.a 1-FNMR) at specific False Acceptance Rate (FARs a.k.a, FMRs) of 1\%.
\paragraph{Baselines}
To benchmark performance on the dataset, we consider several publicly available open-source face recognition models. Although there exists a modality gap between enrollment (RGB) and probe (NIR), this gap is relatively small compared to other heterogeneous settings \cite{george2022prepended}. We evaluate a set of baseline models that are originally trained for standard face recognition tasks, and additionally include models that are explicitly trained on the training split of \dbname.
AdaFace \cite{kim2022adaface} introduced an adaptive margin-based loss for face recognition that uses the feature norm as a proxy for image quality. In our experiments, we used the publicly available AdaFace model with a ResNet-100 backbone trained on the WebFace12M dataset.
LVFace \cite{you2025lvface} introduced a Vision Transformer (ViT) based FR model designed to better exploit large vision models for face recognition. In our experiments, we used the publicly available LVFace model based on a ViT-L backbone trained on the Glint360K dataset.
EdgeFace \cite{george2024edgeface} is a lightweight face recognition model built on a hybrid CNN--Transformer backbone. It is trained on RGB face images from the WebFace dataset \cite{zhu2021webface260m}. In our experiments, we used the publicly available EdgeFace-Base variant.
xEdgeFace \cite{george2025xedgeface} is a contrastive self-distillation framework designed to adapt a pretrained face recognition backbone to heterogeneous face recognition. The method retains the original EdgeFace \cite{george2024edgeface} architecture and introduces cross-modal alignment along with teacher–student supervision to enable effective adaptation while preserving performance on the source (RGB) task. As a result, the model gains robust heterogeneous recognition capability without catastrophic forgetting, while remaining lightweight. 

\paragraph{Experiment details}
For AdaFace, EdgeFace, and LVFace, we use publicly available pretrained models and perform baseline evaluations by comparing each probe sample against all gallery templates using cosine distance. All images are first preprocessed using face detection and alignment, and the resulting $112 \times 112$ crops are fed into the pretrained networks for feature extraction and evaluation. For xEdgeFace, we adopt the EdgeFace backbone and further tune it using the training split of \dbname following the method proposed in \cite{george2025xedgeface}, specifically optimizing it for the cross-spectral setting. The performance reported corresponds to evaluation on the test set.

\begin{table}
\caption{Model performance comparison across different protocols}
\label{tab:combined_results_fr}
\centering
\resizebox{0.49\textwidth}{!}{%
\begin{tabular}{lccccc}
\toprule
\textbf{Protocol} & \textbf{Model} & \textbf{AUC} & \textbf{EER} & \textbf{VR@FAR=1$\%$} & \textbf{Rank-1} \\
\midrule

\multirow{4}{*}{Outdoor}
 & AdaFace   & 99.45 & 2.69 & \textbf{96.01} & 97.42 \\
 & LVFace    & 98.50 & 5.26 & 90.17 & 92.26 \\
 & EdgeFace  & 98.38 & 5.37 & 91.35 & 93.29 \\
 & xEdgeFace & 98.94 & 4.18 & \underline{93.12} & 94.22 \\
\midrule
\multirow{4}{*}{Simulation}
 & AdaFace   & 96.23 & 8.26 & \underline{88.27} & 90.46 \\
 & LVFace    & 95.19 & 11.42 & 79.68 & 83.95 \\
 & EdgeFace  & 94.09 & 11.73 & 82.93 & 84.72 \\
 & xEdgeFace & 96.03 & 8.09 & \textbf{88.63} & 89.68 \\
\midrule
\multirow{4}{*}{Indoor Car}
 & AdaFace   & 98.59 & 3.11 & \textbf{96.24} & 97.11 \\
 & LVFace    & 98.52 & 5.29 & 90.23 & 93.27 \\
 & EdgeFace  & 98.75 & 4.48 & 93.92 & 93.34 \\
 & xEdgeFace & 97.98 & 3.47 & \underline{95.95} & 96.16 \\

\bottomrule
\end{tabular}%
}
\end{table}

\paragraph{Experimental results}
Tab.~\ref{tab:combined_results_fr} summarizes the face recognition performance across the three evaluation protocols. 

\textbf{Outdoor protocol.}
The outdoor protocol achieves the strongest overall performance across all models, indicating that despite the RGB-NIR modality gap, robust FR models perform reasonably well. AdaFace achieves the best performance with an AUC of 99.45\% and the lowest EER of 2.69\%, along with the highest Rank-1 accuracy (97.42\%). This suggests that its adaptive margin mechanism generalizes well even under mild modality shifts. xEdgeFace improves over its backbone (EdgeFace), reducing the EER from 5.37\% to 4.18\% and increasing verification rates, demonstrating the effectiveness of cross-spectral adaptation. 

\textbf{Simulation protocol.}
The simulation protocol is the most challenging scenario, with all models exhibiting a noticeable drop in performance. This can be attributed to the presence of controlled variations such as different glass tint levels and potentially more severe spectral distortions. AdaFace and xEdgeFace achieve comparable performance, with AUC values around 96\% and EERs near 8\%, significantly higher than in other protocols. Importantly, xEdgeFace slightly outperforms AdaFace in VR@FAR=1\%, highlighting the advantage of explicit cross-spectral adaptation under challenging conditions. EdgeFace and LVFace show the largest degradation, indicating that models trained purely on RGB data struggle to generalize in this scenario.

\textbf{Indoor car protocol.}
In the indoor in-vehicle setting, performance remains high but shows a slight degradation compared to outdoor conditions, likely due to more constrained lighting and reflections from vehicle interiors. AdaFace achieves the best overall results in this protocol, with the lowest EER (3.11\%) and highest Rank-1 accuracy (97.11\%), indicating strong robustness to controlled in-car conditions. Notably, xEdgeFace again improves over EdgeFace, confirming that domain adaptation consistently benefits cross-spectral matching. However, the gains are less pronounced than in the simulation scenario, suggesting that the domain gap is smaller in this setting.

Across all protocols, the results highlight two key aspects: (i) the impact of cross-spectral and environmental variations on recognition performance, and (ii) the benefit of adapting models to the target domain. Pretrained RGB models (AdaFace, EdgeFace, and LVFace) demonstrate strong baseline performance, indicating that the RGB–NIR gap in this dataset is small, although it still introduces noticeable degradation under more challenging conditions. Among these, AdaFace consistently achieves the most robust performance, likely due to its quality-aware training strategy. At the same time, the consistent improvements of xEdgeFace over its backbone EdgeFace across all protocols emphasize the importance of domain adaptation for heterogeneous face recognition, with the gains being particularly evident in the more challenging simulation scenario. It should be noted that, EdgeFace and xEdgeFace are significantly more lightweight compared to AdaFace and LVFace, yet they remain competitive, highlighting an important trade-off between efficiency and performance. Overall, while strong pretrained models generalize reasonably well, explicit cross-spectral adaptation is crucial for achieving robust performance under varying operational conditions, especially when illumination changes and sensor differences introduce larger domain shifts.

\begin{figure*}[t]
    \centering
    \includegraphics[width=\linewidth, trim=0 0 0 40, clip]{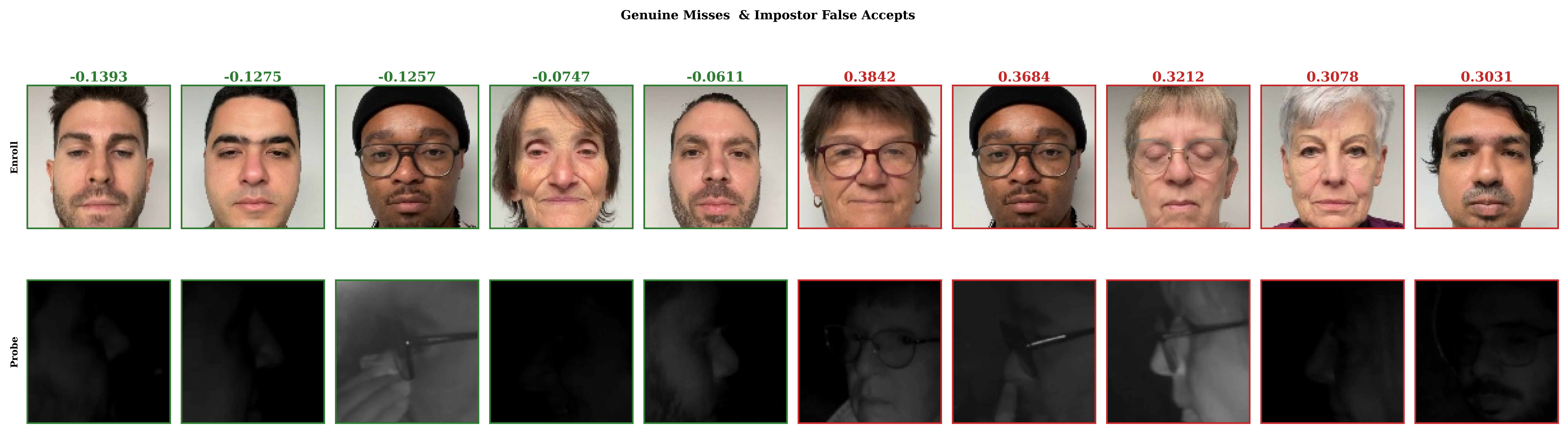}
    \caption{Worst-case verification pairs. \textbf{Left (green):} genuine pairs with the lowest match scores, indicating failure to recognize true matches. \textbf{Right (red):} impostor pairs with the highest match scores, indicating false acceptances. Top row shows enrollment images; bottom row shows corresponding probe images.}
    \label{fig:worst_cases}
\end{figure*}

Figure \ref{fig:worst_cases} shows representative failure cases in the simulation protocol. From the images, it is evident that most failures occur under conditions such as extreme tint, low contrast, extreme profile views, and the presence of occlusions.

\textbf{Effect of Tint Level}
We performed a set of experiments on the controlled subset, reporting AUC separately for each VLT level and for clear glass (Tab. \ref{tab:auc_tint_levels}). Across the evaluated models, performance generally improves as light transmission increases, with clear glass giving the best results. This directly quantifies the effect of tint.

\begin{table}[htbp]
\centering
\caption{AUC values for different FR models across tint levels.}
\label{tab:auc_tint_levels}
\resizebox{0.8\linewidth}{!}{%
\begin{tabular}{lrrrr}
\toprule
\textbf{VLT} &
\textbf{AdaFace} &
\textbf{LVFace} &
\textbf{EdgeFace} &
\textbf{xEdgeFace} \\
\midrule
T05   & 98.36  & 98.72 & 98.42 & 98.20  \\
T15   & 98.48  & 98.09 & 98.12 & 98.60  \\
T20   & 99.31  & 99.04 & 98.23 & 98.72  \\
T35   & 99.62  & 99.14 & 98.75 & 99.35  \\ \hline
clear & 100.00 & 99.86 & 99.87 & 100.00 \\
\bottomrule
\end{tabular}
}
\end{table}

\subsection{Presentation Attack Detection}

As noted in the study in \cite{george2025invisible}, replay attacks do not constitute a significant challenge in the near-infrared (NIR) domain, as they are largely invisible in NIR. Therefore, this work focuses on more challenging and operationally relevant attack types, namely print and mask-based presentation attacks \cite{george2023robust}, which exhibit greater attack potential.

\paragraph{Protocols} We define three evaluation protocols to assess model robustness under different attack conditions: \emph{grandtest}, where all attack types appear across training, development, and evaluation splits, representing a known-attack setting with disjoint subjects and attack instruments; \emph{unseen\_print}, where print attacks are excluded from training and development and appear only at evaluation, simulating an unseen print-attack scenario; and \emph{unseen\_mask}, which follows the same setup for mask attacks to measure generalization to unseen mask-based attacks. A detailed summary of the dataset statistics for each protocol is provided in Table~\ref{tab:pad_protocol_summary}.

\begin{table}[ht]
\centering
\caption{Summary of PAD protocol splits.}
\label{tab:pad_protocol_summary}
\resizebox{0.49\textwidth}{!}{%
\begin{tabular}{llrrrrrr}
\toprule
Protocol & Split & IDs & Frames & Bonafide & Attack & Print & Mask \\
\midrule
\multirow{3}{*}{grandtest}
  & train & 42 & 13\,375 &  9\,856 & 3\,519 & 1\,495 & 2\,024 \\
  & dev   & 14 &  4\,845 &  3\,488 & 1\,357 &    561 &    796 \\
  & eval  & 14 &  5\,501 &  4\,098 & 1\,403 &    664 &    739 \\
\midrule
\multirow{3}{*}{unseen\_print}
  & train & 36 & 11\,843 &  9\,263 & 2\,580 &      0 & 2\,580 \\
  & dev   & 12 &  3\,944 &  2\,965 &    979 &      0 &    979 \\
  & eval  & 22 &  7\,934 &  5\,214 & 2\,720 &  2\,720 &     0 \\
\midrule
\multirow{3}{*}{unseen\_mask}
  & train & 44 & 13\,786 & 11\,546 & 2\,240 &  2\,240 &     0 \\
  & dev   & 14 &  3\,781 &  3\,301 &    480 &    480 &     0 \\
  & eval  & 12 &  6\,154 &  2\,595 & 3\,559 &      0 & 3\,559 \\
\bottomrule
\end{tabular}
}
\end{table}

\paragraph{Metrics} To evaluate the performance of presentation attack detection (PAD) systems, we adopt the standardized metrics defined in ISO/IEC 30107-3 \cite{iso30107-3-2023}. Specifically, we report the Attack Presentation Classification Error Rate (APCER) and the Bona Fide Presentation Classification Error Rate (BPCER), where BPCER corresponds to the rate at which bona fide samples are incorrectly classified as attacks. Additionally, we compute the Average Classification Error Rate (ACER), defined as the mean of APCER and BPCER. The decision threshold is determined on the development set of each protocol using the Equal Error Rate (EER) criterion.

\paragraph{Baselines}
We evaluate several baseline models across the protocols. DeepPixBiS \cite{george2019deep} employs a DenseNet-based architecture with pixel-wise binary supervision to enhance PAD performance. In our implementation, we utilize only the pixel-wise binary loss during training, and final predictions are obtained by averaging the resulting pixel-wise score maps. For CLIP \cite{radford2021learning}, we utilize the vision encoder of the CLIP model, specifically the ViT-B/32 variant. Two training configurations are considered: (i) \emph{CLIP (fc only)}, where only the final fully connected layer is trained while keeping the backbone frozen, and (ii) \emph{CLIP (full)}, where all model parameters are fine-tuned. The DinoV2 \cite{oquab2023dinov2} baseline utilizes the DinoV2 backbone, specifically the ViT-B/14 variant. Only the final fully connected classification layer is fine-tuned, while the backbone remains frozen. ConvNeXtV2-Tiny and EfficientNet-B0 are lightweight convolutional architectures based on ConvNeXt V2 \cite{woo2023convnext} and EfficientNet \cite{tan2019efficientnet}, respectively. Both models are pretrained on ImageNet and subsequently adapted for the binary PAD classification task.
 
\paragraph{Experimental Settings}
All images were first preprocessed with face detection using the SCRFD \cite{guo2021sample} model and alignment, after which they were cropped to a fixed spatial resolution of $224 \times 224$ pixels.
All models were trained for 100 epochs using a learning rate of $1 \times 10^{-4}$ and a weight decay of $1 \times 10^{-6}$, with a batch size of 64. The training process was implemented in \texttt{PyTorch} and used NVIDIA RTX 3090 GPUs. To improve generalization, data augmentation was applied during training, including random horizontal flipping, RandAugment, random rotations up to $30^\circ$, and color jitter with brightness, contrast, and saturation variations of 0.15. To mitigate class imbalance, the training data was sampled in a balanced manner.
    
\paragraph{Experimental Results on PAD}

\begin{table}[ht]
\centering
\caption{Summary of PAD results.}
\label{tab:pad_results}
\resizebox{0.49\textwidth}{!}{%
\begin{tabular}{llrrr}
\toprule
& & \multicolumn{3}{c}{EER} \\
\cmidrule(lr){3-5}
Protocol & Model & APCER & BPCER & ACER \\
\midrule

\multirow{6}{*}{grandtest}
& DeepPixBiS \cite{george2019deep} & 0.00 & 3.30 & 1.60 \\
& CLIP (fc only) & 4.40 & 2.60 & 3.50 \\
& DinoV2 (fc only) & 3.50 & 1.00 & 2.20 \\
& EfficientNet-B0 & 0.10 & 0.90 & \textbf{0.50} \\
& CLIP (full) & 1.60 & 6.50 & 4.00 \\
& ConvNeXtV2-Tiny & 0.30 & 1.00 & \underline{0.70} \\

\midrule
\multirow{6}{*}{unseen\_mask}
& DeepPixBiS \cite{george2019deep} & 91.60 & 0.00 & 45.80 \\
& CLIP (fc only) & 78.80 & 0.00 & 39.40 \\
& DinoV2 (fc only) & 52.30 & 0.10 & \textbf{26.20} \\
& EfficientNet-B0 & 54.50 & 0.00 & \underline{27.20} \\
& CLIP (full) & 83.90 & 1.60 & 42.70 \\
& ConvNeXtV2-Tiny & 73.10 & 0.00 & 36.60 \\

\midrule
\multirow{6}{*}{unseen\_print}
& DeepPixBiS \cite{george2019deep} & 36.70 & 0.00 & \textbf{18.40} \\
& CLIP (fc only) & 67.80 & 0.10 & 34.00 \\
& DinoV2 (fc only) & 53.00 & 0.20 & 26.60 \\
& EfficientNet-B0 & 70.70 & 0.00 & 35.40 \\
& CLIP (full) & 64.70 & 3.40 & 34.00 \\
& ConvNeXtV2-Tiny & 42.80 & 0.10 & \underline{21.50} \\

\bottomrule
\end{tabular}%
}
\end{table}

Table~\ref{tab:pad_results} summarizes the PAD performance across all three protocols. In the grandtest protocol  all models achieve low error rates, with EfficientNet-B0 performing best (ACER = 0.50\%), followed by ConvNeXtV2-Tiny (0.70\%). This indicates that the evaluated models perform well in the known-attack setting. CLIP-based models show comparatively higher errors, particularly in BPCER.

In contrast, performance drops sharply in the unseen attack scenarios. For unseen\_mask, all models exhibit very high APCER, with DinoV2 (fc only) performing best (ACER = 26.20\%), suggesting better generalization from self-supervised (SSL) features. However, overall results remain poor, highlighting the difficulty of mask attack detection under domain shift. For unseen\_print, the degradation is less severe. DeepPixBiS achieves the best result (ACER = 18.40\%), followed by ConvNeXtV2-Tiny (21.50\%).

These results motivate further research on domain adaptation and self-supervised representations for generalization to unseen attacks.

\section{Discussion}

The experiments on \dbname demonstrate that both FR and PAD remain challenging in realistic vehicular border-control conditions, even when strong pretrained models are used. On the FR side, the results show that the RGB-NIR modality gap is manageable but still non-negligible, especially under more difficult conditions such as the simulation protocol, where glass tint and controlled illumination changes introduce a larger domain shift. AdaFace consistently provides the strongest overall baseline, while xEdgeFace shows clear gains over EdgeFace across all protocols, confirming the value of explicit cross-spectral adaptation. Importantly, EdgeFace and xEdgeFace are significantly smaller and more lightweight than AdaFace and LVFace, yet they remain competitive, making them particularly attractive for deployment in practical resource-constrained systems. On the PAD evaluations, the grandtest protocol results indicate that known attacks can be detected reliably, but performance degrades substantially in unseen attack settings, especially for mask attacks, revealing a clear generalization gap. These findings suggest that \dbname is a challenging and realistic benchmark that exposes limitations not only in cross-spectral face recognition, but also in robust spoof detection under operational conditions, highlighting the need for models that jointly address efficiency, domain shift, and generalization to unseen scenarios.

\section{Conclusion}
In this work, we introduced the new \dbname and \dbnamepad datasets for cross-spectral through-glass face analysis in on-the-move vehicular border-control scenarios, together with benchmark protocols for both face recognition and presentation attack detection. The datasets capture realistic operational challenges, including RGB-to-NIR matching, through-glass acquisition, unconstrained viewpoints, varying illumination, and presentation attacks. Experimental results show that while modern pretrained face recognition models already provide strong baselines, their performance still drops under more challenging environmental and spectral conditions, and explicit cross-spectral adaptation further improves robustness. Similarly, PAD results reveal that detecting known attacks is relatively easy, whereas generalization to unseen attacks remains difficult. Overall, \dbname and \dbnamepad provide a valuable benchmark for advancing research on robust and efficient biometric systems for real-world border-control applications. To support reproducibility and further research, the dataset, evaluation protocols, and code are publicly available.
\section{Ethics Statement}

All participants in this data collection volunteered and provided informed consent, agreeing to the collection and use of their data for the specified research purposes. The project under which the data was collected was approved by the institution’s Data Research Ethics Committee (DREC).

\section{Acknowledgments}

The project leading to this work has received funding from Frontex under the Frontex Research Grants Programme.
Call for Proposals 2024/CFP/INNOVATE/01 Grant Agreement No. 2025/280. This work reflects only the authors’ view. Neither the European Union nor Frontex are responsible for any use that may be made of the information it contains.
This research was also partly funded by the European Union project CarMen (Grant Agreement No. 101168325).

{\small
\bibliographystyle{ieee}
\bibliography{egbib}
}

\end{document}